\documentclass[10pt,twocolumn,letterpaper]{article}

\usepackage{cvpr}              
\definecolor{cvprblue}{rgb}{0.21,0.49,0.74}
\usepackage[pagebackref,breaklinks,colorlinks,allcolors=cvprblue]{hyperref}
\usepackage{booktabs}
\usepackage{multirow}
\usepackage[table]{xcolor} 
\usepackage{adjustbox} 
\usepackage[T1]{fontenc}
\usepackage[utf8]{inputenc}
\def\paperID{*****} 
\def\confName{CVPR}
\def\confYear{2026}

\title{
\raisebox{0.01em}{\includegraphics[height=1.2em]{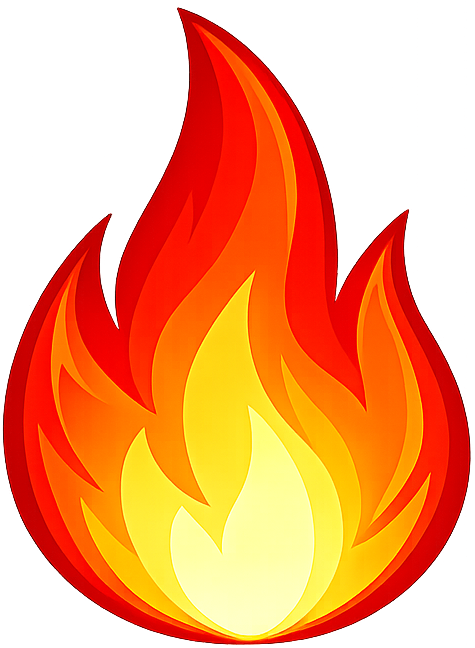}}
HOT: Harmonic-Constrained Optimal Transport \\
for Remote Photoplethysmography Domain Adaptation
}

\author{
{\large
Ba-Thinh Nguyen$^{1}$,
Thi-Duyen Ngo$^{1}$,
Thanh-Trung Huynh$^{3}$,
Thanh-Ha Le$^{1,\dagger}$,
Huy-Hieu Pham$^{2,3,4,\dagger}$
}\\[0.6em]
$^{1}$VNU University of Engineering and Technology, Hanoi, Vietnam\\
$^{2}$VinUni-Illinois Smart Health Center, VinUniversity, Hanoi, Vietnam\\
$^{3}$College of Engineering and Computer Science, VinUniversity, Hanoi, Vietnam\\
$^{4}$Center for Innovations in Health Sciences, VinUniversity, Hanoi, Vietnam\\
[0.4em]
{\tt\small
\{22028163, duyennt, ltha\}@vnu.edu.vn,
\{trung.ht, hieu.ph\}@vinuni.edu.vn
}
}

\begin{document}
\maketitle

\begingroup
\renewcommand{\thefootnote}{\fnsymbol{footnote}}
\setcounter{footnote}{0}
\footnotetext[2]{Corresponding authors: ltha@vnu.edu.vn; hieu.ph@vinuni.edu.vn}
\endgroup
\begin{abstract}
Remote photoplethysmography (rPPG) enables non-contact physiological measurement from facial videos; however, its practical deployment is often hindered by substantial performance degradation under domain shift. While recent deep learning–based rPPG methods have achieved strong performance on individual datasets, they frequently overfit to appearance-related factors—such as illumination, camera characteristics, and color response—that vary significantly across domains. To address this limitation, we introduce frequency domain adaptation (FDA) as a principled strategy for modeling appearance variation in rPPG. By transferring low-frequency spectral components that encode domain-dependent appearance characteristics, FDA encourages rPPG models to learn invariance to appearance variations while retaining cardiac-induced signals. To further support physiologically consistent alignment under such appearance variation, we propose Harmonic-Constrained Optimal Transport (HOT), which leverages the harmonic property of cardiac signals to guide alignment between original and FDA-transferred representations. Extensive cross-dataset experiments demonstrate that the proposed FDA and HOT framework effectively enhances the robustness and generalization of rPPG models across diverse datasets.
\end{abstract}    
\section{Introduction}
\label{sec:intro}

Remote photoplethysmography (rPPG) is a non-contact technique for estimating physiological signals, such as blood volume pulse (BVP) and heart rate, from facial videos. By exploiting subtle color variations caused by cardiac-induced blood volume changes, rPPG enables convenient and unobtrusive physiological monitoring using consumer-grade cameras. Owing to its practicality and low deployment cost, rPPG has attracted increasing interest in healthcare monitoring, affective computing, and human--computer interaction~\cite{huang2023challenges}.

\begin{figure}[t]
  \centering
  \includegraphics[width=\linewidth]{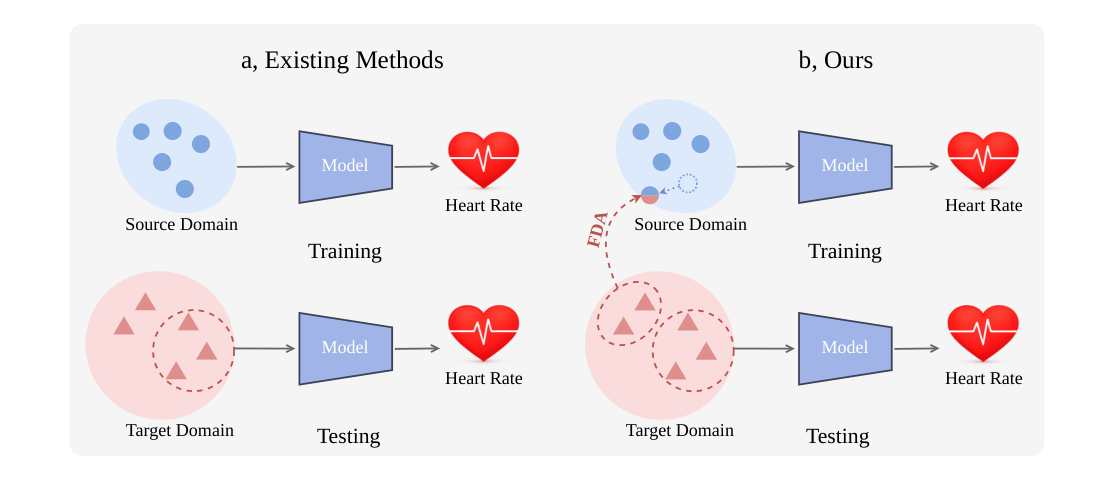}
  \caption{Comparison with existing rPPG methods. 
  Conventional approaches are trained on labeled source-domain data and directly deployed to the target domain, resulting in performance degradation under domain shift. 
  In contrast, ours leverages unlabeled target-domain samples to synthesize source data with target-domain appearance during training, encouraging appearance-invariant learning and improving cross-domain generalization.}
  \label{fig:intro_comparison}
\end{figure}

Early rPPG studies mainly relied on traditional signal processing pipelines, including blind source separation and skin reflection models, to extract pulse-related information from RGB signals~\cite{verkruysse2008remote, poh2010non, de2013robust, wang2016algorithmic}. Representative methods exploit handcrafted assumptions on color decorrelation or optical skin properties, allowing reasonable performance under controlled conditions. However, these approaches are highly sensitive to noise, motion, and illumination variations, limiting their applicability in real-world scenarios.

With the development of deep learning, end-to-end rPPG models have achieved substantial performance gains by learning spatio-temporal representations directly from video data. Convolutional and transformer-based architectures can suppress certain types of noise and capture temporal dependencies, achieving state-of-the-art results on several benchmark datasets~\cite{chen2018deepphys, yu2019remote, liu2020multi, yu2022physformer, liu2023efficientphys, zou2025rhythmformer, nguyen2025reperio}. Nevertheless, most existing deep rPPG methods are trained and evaluated on a single dataset, implicitly assuming consistent data distributions between training and testing.

In practice, rPPG performance often degrades significantly when models are applied across datasets or environments due to domain shift caused by changes in illumination, camera characteristics, recording protocols, and subject appearance. Despite its importance, this problem remains underexplored in the rPPG literature. In particular, unlike other vision tasks, frequency domain adaptation (FDA)~\cite{yang2020fda}, which transfers appearance statistics by modifying low-frequency spectral components while preserving semantic content, has not yet been investigated for rPPG. This is non-trivial because rPPG relies on extremely subtle color variations, and na\"ively applying appearance transformations may corrupt physiological information.
 
In this work, we present the first study introducing FDA for domain generalization in rPPG. By leveraging FDA, we generate appearance-shifted video samples that emulate realistic cross-domain variations while preserving the underlying physiological dynamics. However, enforcing consistency between original and FDA-transferred samples solely through embedding similarity may yield geometrically consistent yet physiologically incompatible correspondences. To address this issue, we formulate temporal alignment as an optimal transport problem and introduce Harmonic-Constrained Optimal Transport (HOT). Specifically, HOT augments the transport cost with a harmonic regularization term derived from the spectral structure of cardiac signals, biasing the transport plan toward physiologically coherent token correspondences under appearance variation.

Overall, our contributions can be summarized as follows:

1. We present an FDA-based framework for rPPG domain adaptation, using unlabeled target data to generate appearance-shifted training videos that better mimic target-domain variations.

2. We propose HOT, which imposes a harmonic constraint on Sinkhorn transport to encourage physiologically consistent alignment between original and FDA-transferred representations.

3. Extensive cross-dataset experiments and ablations show that FDA + HOT consistently improves cross-domain rPPG estimation performance.

The rest of this paper is organized as follows. Section~\ref{sec:relatedwork} reviews related work on rPPG estimation and cross-domain robustness. Section~\ref{sec:method} presents the proposed HOT framework, including FDA and HOT. Section~\ref{sec:experiments} describes the experimental setup and reports the main cross-dataset results, while Section~\ref{sec:ablation_study} provides ablation and sensitivity analyses. Finally, Section~\ref{sec:Conclusion} concludes the paper and discusses limitations and future directions.
\section{Related Work}
\label{sec:relatedwork}
\begin{figure*}[hbt] 
    \centering
    \includegraphics[width=0.8\linewidth]{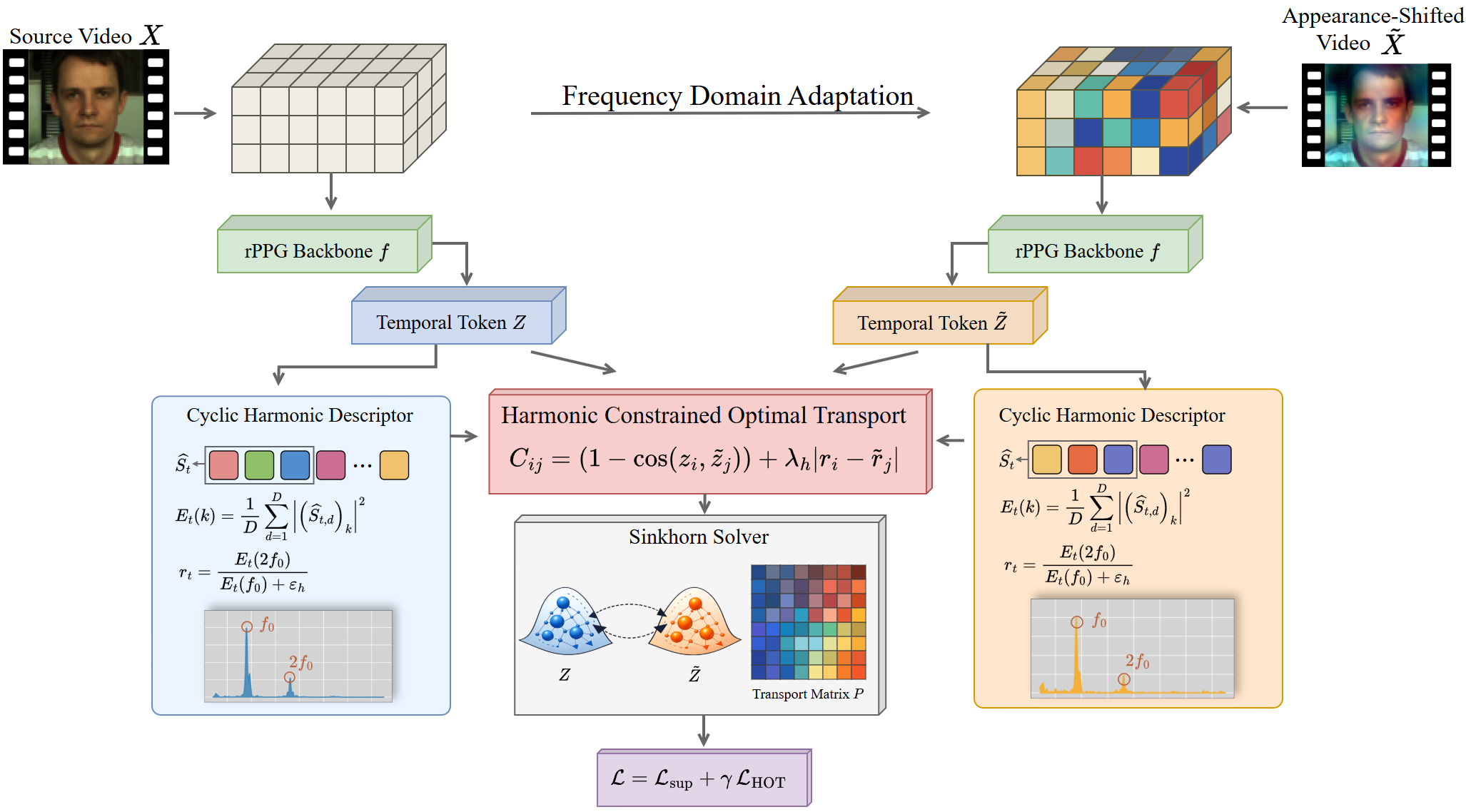} 

    \caption{Overall architecture of HOT.}

    \label{fig:HOT Architecture} 
\end{figure*}
\subsection{Traditional rPPG Methods}
Early studies on remote photoplethysmography mainly relied on handcrafted signal processing techniques to extract physiological signals from facial videos. Representative approaches employ blind source separation methods, such as independent component analysis (ICA)~\cite{poh2010non} and principal component analysis (PCA)~\cite{lewandowska2012measuring}, to decompose observed color signals into pulse-related and noise components. Other methods leverage skin reflection models and color space transformations to isolate chrominance signals induced by cardiac activity, including CHROM~\cite{de2013robust}, POS~\cite{wang2016algorithmic}, and their variants. Under stable illumination and limited motion, these traditional pipelines can achieve reasonable accuracy without learning-based models. However, their strong reliance on fixed assumptions and heuristic designs makes them highly sensitive to motion artifacts, illumination changes, and camera-dependent noise, which significantly limits their robustness in unconstrained scenarios.

\subsection{Deep Learning-based rPPG Estimation}
Driven by the success of deep learning in video understanding, recent rPPG research has increasingly focused on data-driven approaches that learn spatio-temporal representations directly from video inputs. Early deep models exploit normalized difference frames and attention mechanisms to enhance subtle pulse-related variations~\cite{yu2019remote, liu2020multi}. Subsequent works adopt convolutional architectures with temporal modeling modules, such as temporal shift operations and multi-scale feature fusion, to improve efficiency and robustness~\cite{li2023channel, liu2023efficientphys}. Transformer-based models further extend this line of research by capturing long-range temporal dependencies and leveraging self-attention mechanisms for rPPG estimation~\cite{yu2022physformer, zou2025rhythmformer, qian2024dual, mvrd, physllm}.

In addition to architectural advances, several studies incorporate physiological priors, including periodicity and frequency-domain constraints, into network design or training objectives. Methods based on spatio-temporal maps, masked pretraining, and self-supervised learning have also been proposed to reduce annotation dependency and improve representation quality~\cite{liu2024rppg, zhang2024maskfusionnet, savic2025rs+, qian2025physdiff}. While these deep learning–based approaches achieve strong performance on individual datasets, they are typically trained and evaluated under dataset-specific conditions and assume consistent data distributions between training and testing.

\subsection{Robustness and Cross-Domain rPPG Learning}
Despite the progress of deep rPPG models, their performance often degrades when applied across datasets or environments due to domain shift caused by variations in illumination, camera characteristics, recording protocols, and subject appearance. Several recent works have attempted to improve robustness by enhancing temporal modeling, exploiting periodicity, or learning disentangled representations that separate physiological signals from noise~\cite{liu2023robust, jiang2025lsts, nguyen2025reperio}. Other approaches explore domain generalization or test-time adaptation strategies to mitigate cross-domain degradation without requiring labeled target data~\cite{lu2021dual, du2023dual, lee2025continual}.

Nevertheless, most existing methods address domain shift implicitly through architectural design or feature regularization, rather than via explicit data-level domain transformation (see Fig.~\ref{fig:intro_comparison}). In particular, although frequency-domain techniques have been widely studied for image domain adaptation in general vision tasks, their use in rPPG remains unexplored. Moreover, existing rPPG methods rarely consider explicit alignment between original and appearance-shifted samples under physiological constraints. These gaps motivate our work, which introduces FDA to rPPG and further employs HOT to achieve physiologically consistent alignment under domain shift.

\section{Methodology}
\label{sec:method}

We propose \textbf{HOT} (see Fig.~\ref{fig:HOT Architecture}), a physiology-guided temporal alignment framework for rPPG domain adaptation. Given a source-domain clip and its FDA-generated appearance-shifted counterpart, HOT aligns their temporal features through an optimal transport objective constrained by local harmonic statistics.

\subsection{Frequency-Domain Adaptation}

Let $\mathbf{X}\in\mathbb{R}^{B\times C\times T\times H\times W}$ denote a batch of source-domain video clips.
For each clip, let $x_t\in\mathbb{R}^{C\times H\times W}$ be the $t$-th frame and let $x_{r}\in\mathbb{R}^{C\times H\times W}$ be a reference frame sampled from the target domain.
We apply a channel-wise 2D discrete Fourier transform over the spatial dimensions:
\begin{equation}
\left[\mathcal{F}\{x\}\right]_{u,v}
=
\left[\mathcal{A}\{x\}\right]_{u,v}
\, e^{\,i\left[\Phi\{x\}\right]_{u,v}}.
\end{equation}
where $\mathcal{A}\{x\}=|\mathcal{F}\{x\}|$ and $\Phi\{x\}=\angle\!\bigl(\mathcal{F}\{x\}\bigr)$ denote the amplitude and phase spectra, respectively, and $(u,v)$ denotes the spatial-frequency index. 

\begin{figure}[t]
  \centering
  \includegraphics[width=\linewidth]{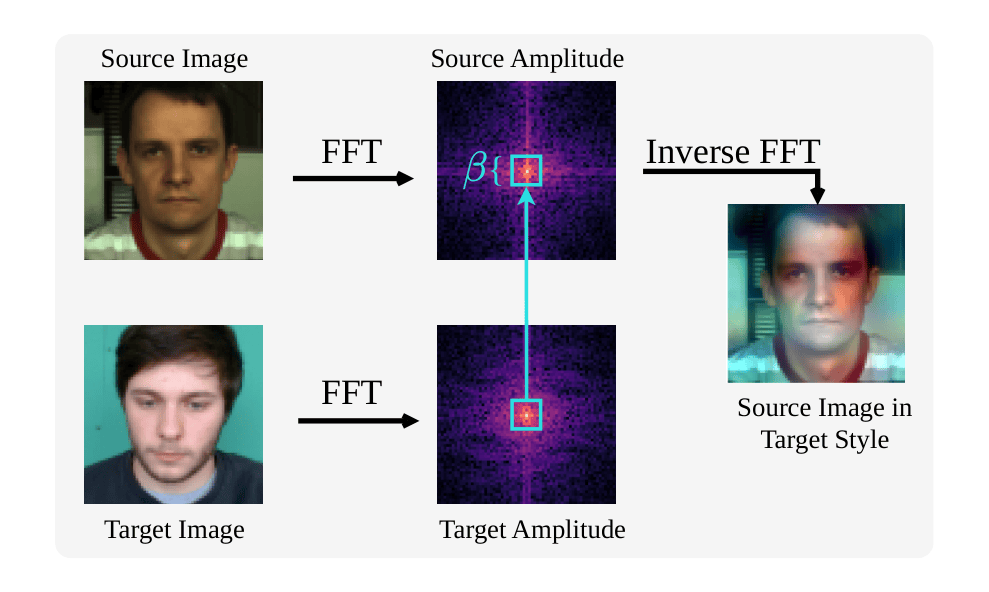}
    \caption{Illustration of FDA: low-frequency amplitude replacement in the Fourier domain while preserving the phase spectrum.}
  \label{fig:FDA}
\end{figure}

Let $\Omega_\beta$ be a centered low-frequency region determined by $\beta\in(0,1)$. 
Define the binary mask
\begin{equation}
M_\beta(u,v)=\mathbb{I}\big[(u,v)\in\Omega_\beta\big],
\end{equation}
where $\mathbb{I}[\cdot]$ denotes the indicator function, which equals $1$ if the condition is true and $0$ otherwise. Let $\odot$ denote the Hadamard product. FDA then constructs an adapted amplitude spectrum by replacing only the low-frequency amplitudes of $x_t$ with those of $x_{r}$ (see Fig.~\ref{fig:FDA}):
\begin{equation}
\mathcal{A}_{\mathrm{FDA}}(x_t,x_{r})
=
\big(\mathbf{1}-M_\beta\big)\odot \mathcal{A}(x_t)
+
M_\beta\odot \mathcal{A}(x_{r}).
\end{equation}
The stylized frame $\tilde{x}_t$ is then obtained by combining the adapted amplitude with the original phase of $x_t$ and applying the inverse transform:
\begin{equation}
\tilde{x}_t
=
\mathcal{F}^{-1}\!\left(
\mathcal{A}_{\mathrm{FDA}}(x_t,x_{r})\odot
e^{i\,\Phi(x_t)}
\right).
\end{equation}
Applying the above operation independently to all frames yields the stylized clip
\begin{equation}
\tilde{\mathbf{X}}=\{\tilde{x}_t\}_{t=1}^{T}.
\end{equation}
Since domain discrepancy in rPPG datasets primarily arises from camera spectral response and illumination variations, which are largely reflected in the DC and low-frequency amplitude spectrum, altering only this spectral band enables FDA to correct domain-specific photometric bias while preserving the original phase spectrum.
Consequently, the resulting appearance adaptation reduces cross-domain gaps without perturbing subtle rPPG-related dynamics.

\subsection{Cyclic Local Harmonic Descriptor}

An rPPG backbone network $f$ maps $\mathbf{X}$ to intermediate spatio-temporal feature maps
\begin{equation}
\mathbf{F} = f(\mathbf{X}) \in \mathbb{R}^{B \times D \times T' \times H' \times W'},
\end{equation}
where $D$ denotes the feature dimension, $T'$ the temporally downsampled length, and $H' \times W'$ the spatial resolution of the feature map.
We omit the batch index below for clarity.

\paragraph{Temporal Tokenization.}
We first aggregate spatial information at each time index into a single temporal token.
For $t = 1,\dots,T'$, we define
\begin{equation}
z_t = \frac{1}{H'W'} 
\sum_{h=1}^{H'} \sum_{w=1}^{W'} \mathbf{F}_{:,t,h,w}
\;\in\; \mathbb{R}^{D},
\end{equation}
which yields the token sequence \(\mathbf{Z}=\{z_t\}_{t=1}^{T'}\). Applying the same process to the stylized clip \(\tilde{\mathbf{X}}\) yields \(\tilde{\mathbf{Z}}=\{\tilde{z}_t\}_{t=1}^{T'}\).

\paragraph{Cyclic Local Temporal Signal.}
To characterize local physiological dynamics, we associate each token $z_t$ with a cyclic temporal neighborhood of fixed length $W$.
We define an ordered cyclic index mapping
\begin{equation}
\pi_t(m)=\bigl((t+m-1)\bmod T'\bigr)+1,\qquad m=0,\dots,W-1,
\end{equation}
and construct the local temporal signal
\begin{equation}
\mathbf{S}_t=\bigl[z_{\pi_t(m)}\bigr]_{m=0}^{W-1}\in\mathbb{R}^{W\times D},
\end{equation}
where the row associated with offset $m$ is given by the token indexed by $\pi_t(m)$.
This cyclic construction guarantees a valid neighborhood for every $t$, including boundary positions, which is particularly important for short rPPG clips.

\paragraph{Harmonic Statistic.}
We extract a frequency-domain descriptor from each local signal $\mathbf{S}_t$ along the temporal dimension.
Let $\mathbf{S}_t(m,d)$ denote the entry at row $m\in\{0,\dots,W-1\}$ and channel $d\in\{1,\dots,D\}$.
We first apply a Hann window to reduce spectral leakage:
\begin{equation}
w(m)=0.5-0.5\cos\!\left(\frac{2\pi m}{W-1}\right),\qquad m=0,\dots,W-1,
\end{equation}
and obtain the windowed signal
\begin{equation}
\widetilde{\mathbf{S}}_t(m,d)=w(m)\,\mathbf{S}_t(m,d).
\end{equation}
For each channel $d$, we compute the discrete Fourier transform along the temporal axis:
\begin{equation}
\widehat{\mathbf{S}}_{t,d}(k)
=
\sum_{m=0}^{W-1}\widetilde{\mathbf{S}}_t(m,d)\,e^{-2\pi i k m/W},
\qquad k=0,\dots,K-1.
\end{equation}
where
\begin{equation}
K = \left\lfloor \frac{W}{2} \right\rfloor + 1
\end{equation}
denotes the number of unique frequency bins for a real-valued signal.
We then define the channel-averaged spectral energy as
\begin{equation}
E_t(k) 
= \frac{1}{D} \sum_{d=1}^{D} 
  \big|\widehat{\mathbf{S}}_{t,d}(k)\big|^2,
\quad k = 0,\dots,K-1.
\end{equation}
Let $f_k$ denote the frequency corresponding to bin $k$ (with resolution $\Delta f$), and 
\begin{equation}
\mathcal{K} = \big\{ k \;\big|\; f_k \in [f_{\min}, f_{\max}] \big\}
\end{equation}
be the set of indices whose frequencies fall within the physiological band (e.g., heart-rate band).
The dominant frequency index in this band is
\begin{equation}
k_1(t) = \arg\max_{k \in \mathcal{K}} E_t(k),
\end{equation}
and the corresponding second harmonic index is defined as
\begin{equation}
k_2(t) = \min\big( 2 k_1(t), \, K-1 \big).
\end{equation}
We finally define the local harmonic ratio
\begin{equation}
r_t 
= \frac{E_t\big(k_2(t)\big)}
       {E_t\big(k_1(t)\big) + \varepsilon_h},
\quad \varepsilon_h > 0,
\end{equation}
which yields the harmonic descriptor sequence
\begin{equation}
\mathbf{r} = (r_1,\dots,r_{T'}).
\end{equation}
An analogous sequence $\tilde{\mathbf{r}} = (\tilde{r}_1,\dots,\tilde{r}_{T'})$ is computed from the stylized token sequence $\tilde{\mathbf{Z}}$.

\subsection{Harmonic-Constrained Optimal Transport}

Given the source and stylized temporal token sequences $\mathbf{Z}$ and $\tilde{\mathbf{Z}}$, together with their harmonic descriptors $\mathbf{r}$ and $\tilde{\mathbf{r}}$, we align them via an optimal transport objective that penalizes both geometric and harmonic discrepancies.

For tokens $z_i$ and $\tilde{z}_j$, we define the transport cost
\begin{equation}
C_{ij} 
= \big( 1 - \cos(z_i, \tilde{z}_j) \big)
  + \lambda_h \, \big| r_i - \tilde{r}_j \big|,
\end{equation}
where $\cos(\cdot,\cdot)$ denotes the cosine similarity and $\lambda_h > 0$ balances feature-level dissimilarity and harmonic inconsistency.

Let $\mathbf{a}, \mathbf{b} \in \Delta^{T'}$ denote marginal distributions over the source and stylized tokens, respectively.
In our experiments, we use uniform marginals, i.e., $a_i = b_j = 1/T'$.
The harmonic-constrained optimal transport loss is defined as
\begin{equation}
\mathcal{L}_{\text{HOT}} =
\min_{\mathbf{P} \in \mathcal{U}(\mathbf{a}, \mathbf{b})}
\sum_{i=1}^{T'} \sum_{j=1}^{T'} P_{ij} C_{ij},
\end{equation}
where the admissible transport plans satisfy
\begin{equation}
\mathcal{U}(\mathbf{a}, \mathbf{b}) =
\big\{ \mathbf{P} \in \mathbb{R}_+^{T' \times T'} 
\;\big|\; 
\mathbf{P}\mathbf{1} = \mathbf{a}, \,
\mathbf{P}^\top \mathbf{1} = \mathbf{b} \big\}.
\end{equation}
To obtain a numerically stable and efficient approximation, we employ an entropy-regularized formulation and solve it using the Sinkhorn~\cite{cuturi2013sinkhorn} algorithm in the log-domain.

\subsection{Training Objective}

Let $\mathbf{y}\in\mathbb{R}^{T}$ denote the ground-truth rPPG signal and let $\hat{\mathbf{y}}\in\mathbb{R}^{T}$ be the predicted rPPG signal obtained by applying an rPPG regression head on top of the backbone features extracted from the input clip $\mathbf{X}$.
Following prior work, we adopt the negative Pearson correlation loss as the supervised objective:
\begin{equation}
\mathcal{L}_{\text{sup}}
= 1 - \rho(\mathbf{y},\hat{\mathbf{y}})
= 1 - \frac{\mathrm{cov}(\mathbf{y},\hat{\mathbf{y}})}
{\sqrt{\mathrm{cov}(\mathbf{y},\mathbf{y})\;\mathrm{cov}(\hat{\mathbf{y}},\hat{\mathbf{y}})}},
\label{eq:neg_pearson}
\end{equation}
where $\mathrm{cov}(\cdot,\cdot)$ denotes the covariance function and $\rho(\mathbf{y},\hat{\mathbf{y}})$ is the Pearson correlation coefficient between the two signals.

The final training objective combines the supervised loss and the harmonic-constrained alignment loss:
\begin{equation}
\mathcal{L} = \mathcal{L}_{\text{sup}} + \gamma \, \mathcal{L}_{\text{HOT}},
\label{eq:final_obj}
\end{equation}
where $\gamma > 0$ controls the contribution of the HOT term.
This objective encourages the model to produce rPPG estimates that are both accurate with respect to ground-truth signals and harmonically consistent across appearance-shifted views of the same underlying physiology.

\begin{table*}[t]
\centering
\footnotesize
\setlength{\tabcolsep}{3.5pt}
\renewcommand{\arraystretch}{1.15}
\newcommand{\NA}{--}
\resizebox{\linewidth}{!}{%
\begin{tabular}{ll*{16}{c}}
\toprule
\multirow{2}{*}{Model} & \multirow{2}{*}{Venue} &
\multicolumn{4}{c}{PURE $\rightarrow$ UBFC-rPPG} &
\multicolumn{4}{c}{UBFC-rPPG $\rightarrow$ PURE} &
\multicolumn{4}{c}{PURE $\rightarrow$ MMPD} &
\multicolumn{4}{c}{UBFC-rPPG $\rightarrow$ MMPD} \\
\cmidrule(lr){3-6}\cmidrule(lr){7-10}\cmidrule(lr){11-14}\cmidrule(lr){15-18}
& 
& MAE$\downarrow$ & MAPE$\downarrow$ & RMSE$\downarrow$ & $r\uparrow$
& MAE$\downarrow$ & MAPE$\downarrow$ & RMSE$\downarrow$ & $r\uparrow$
& MAE$\downarrow$ & MAPE$\downarrow$ & RMSE$\downarrow$ & $r\uparrow$
& MAE$\downarrow$ & MAPE$\downarrow$ & RMSE$\downarrow$ & $r\uparrow$ \\
\midrule

DeepPhys~\cite{chen2018deepphys}
& ECCV'18
& 1.46 & 1.65 & 3.25 & 0.98
& 7.63 & 7.35 & 21.97 & 0.67
& 22.09 & 24.38 & 30.38 & 0.04
& 24.50 & 27.29 & 31.79 & 0.01 \\

\rowcolor{gray!25}\quad + HOT
& \NA
& \textbf{1.31} & \textbf{1.48} & \textbf{3.04} & \textbf{0.99}
& \textbf{7.46} & \underline{7.48} & \textbf{21.95} & 0.67
& \textbf{19.73} & \textbf{20.07} & \textbf{26.46} & \textbf{0.08}
& \textbf{20.65} & \textbf{22.84} & \textbf{24.97} & \textbf{0.05} \\

PhysNet~\cite{yu2019remote}
& BMVC'19
& 1.70 & 1.23 & 3.63 & 0.99
& 7.87 & 15.89 & 18.29 & 0.81
& 16.06 & 18.46 & 23.59 & 0.10
& 11.25 & 18.00 & 23.59 & 0.34 \\

\rowcolor{gray!25}\quad + HOT
& \NA
& \textbf{1.17} & \underline{1.26} & \textbf{3.10} & 0.99
& \textbf{5.38} & \textbf{12.61} & \textbf{16.52} & \textbf{0.87}
& \textbf{12.87} & \textbf{15.90} & \textbf{19.79} & \textbf{0.16}
& \textbf{10.36} & \textbf{14.08} & \textbf{18.42} & \textbf{0.38} \\

TSCAN~\cite{liu2020multi}
& NeurIPS'20
& 1.32 & 1.49 & 3.01 & 0.98
& 5.98 & 6.65 & 19.68 & 0.73
& 18.55 & 20.20 & 27.13 & 0.06
& 18.13 & 19.92 & 26.78 & 0.05 \\

\rowcolor{gray!25}\quad + HOT
& \NA
& \textbf{1.17} & \textbf{1.31} & \textbf{2.81} & \textbf{0.99}
& \textbf{5.03} & \textbf{5.76} & \textbf{17.97} & \textbf{0.79}
& \textbf{15.39} & \textbf{16.21} & \textbf{24.92} & \textbf{0.11}
& \textbf{14.97} & \textbf{14.60} & \textbf{20.95} & \textbf{0.07} \\

PhysFormer~\cite{yu2022physformer}
& CVPR'22
& 3.08 & 3.10 & 9.66 & 0.85
& 10.66 & 20.43 & 21.51 & 0.72
& 19.66 & 22.09 & 26.38 & 0.05
& 18.29 & 23.31 & 25.49 & 0.07 \\

\rowcolor{gray!25}\quad + HOT
& \NA
& \textbf{2.92} & \textbf{2.63} & \textbf{7.61} & \textbf{0.89}
& \textbf{7.52} & \textbf{16.42} & \textbf{17.25} & \textbf{0.79}
& \textbf{18.10} & \textbf{20.61} & \textbf{23.51} & \textbf{0.07}
& \textbf{15.08} & \textbf{16.63} & \textbf{20.13} & \textbf{0.12} \\

EfficientPhys~\cite{liu2023efficientphys}
& WACV'23
& 1.38 & 1.54 & 2.99 & 0.98
& 4.14 & 4.00 & 13.68 & 0.88
& 19.84 & 21.70 & 28.24 & 0.03
& 16.94 & 18.38 & 25.98 & 0.09 \\

\rowcolor{gray!25}\quad + HOT
& \NA
& \textbf{1.06} & \textbf{1.39} & \textbf{2.75} & \textbf{0.99}
& \textbf{3.82} & \textbf{3.90} & \textbf{12.52} & 0.88
& \textbf{17.91} & \textbf{20.57} & \textbf{27.89} & \textbf{0.05}
& \textbf{13.02} & \textbf{15.95} & \textbf{22.13} & \textbf{0.12} \\

RhythmFormer~\cite{zou2025rhythmformer}
& PR'25
& 2.72 & 2.90 & 6.31 & 0.93
& 8.02 & 15.71 & 18.14 & 0.74
& 17.53 & 20.20 & 25.51 & 0.18
& 14.95 & 20.46 & 21.16 & 0.18 \\

\rowcolor{gray!25}\quad + HOT
& \NA
& \textbf{1.84} & \textbf{2.62} & \textbf{5.02} & \textbf{0.96}
& \textbf{6.94} & \textbf{10.53} & \textbf{15.26} & \textbf{0.76}
& \textbf{14.37} & \textbf{19.33} & \textbf{23.02} & \textbf{0.19}
& \textbf{12.15} & \textbf{17.83} & \textbf{18.94} & \textbf{0.21} \\

\bottomrule
\end{tabular}}
\caption{Cross-dataset test results. PURE$\rightarrow$UBFC-rPPG means the model is trained on PURE and tested on UBFC-rPPG. Lower is better for MAE/MAPE/RMSE, while higher is better for Pearson correlation $r$.}
\label{tab:main_comparison}
\end{table*}

\begin{figure*}[hbt] 
    \centering
    \includegraphics[width=0.8\linewidth]{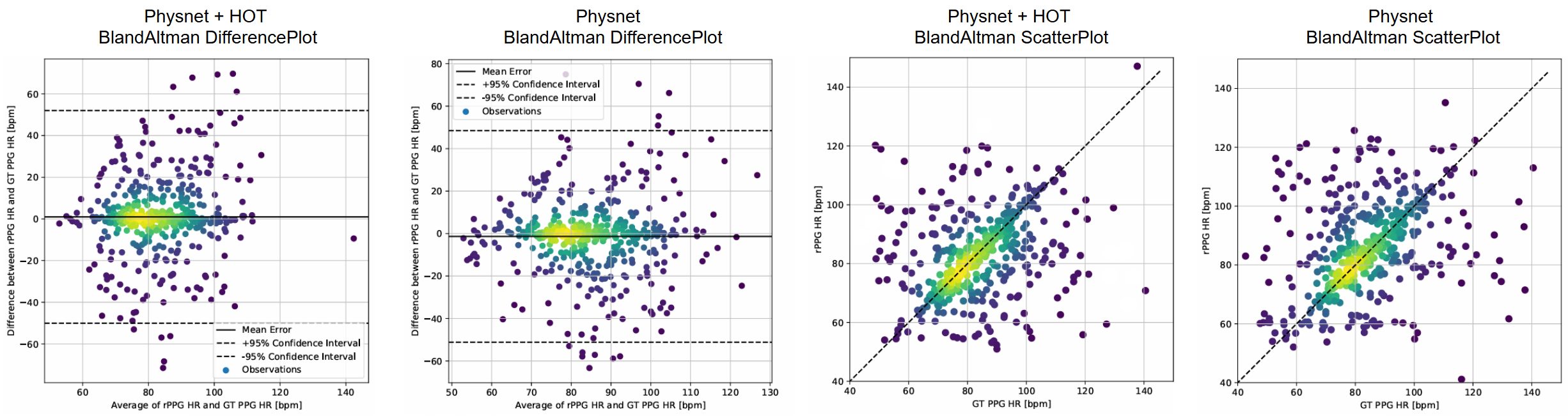} 

    \caption{Cross-dataset evaluation on UBFC-rPPG $\rightarrow$ MMPD. From left to right: Bland--Altman plot of PhysNet + HOT, Bland--Altman plot of PhysNet, scatter plot of PhysNet + HOT, and scatter plot of PhysNet. The dashed horizontal lines in the Bland--Altman plots denote the 95\% limits of agreement, while the dashed diagonal lines in the scatter plots denote the identity line.}

    \label{fig:BlandAltman} 
\end{figure*}

\section{Experiments}
\label{sec:experiments}

\subsection{Datasets}
We evaluate our method on three widely used rPPG datasets: PURE~\cite{stricker2014non}, UBFC-rPPG~\cite{bobbia2019unsupervised}, and MMPD~\cite{tang2023mmpd}.

\paragraph{PURE} is a high-quality rPPG benchmark consisting of 60 video recordings from 10 subjects performing six controlled head-motion activities, including sitting still, talking, slow head movement, quick head movement, small head rotation, and medium head rotation. Videos are recorded at a resolution of 640×480 pixels and 30 fps using lossless PNG encoding, and synchronized with ground-truth PPG signals captured by a CMS50E pulse oximeter at 60 Hz. Due to its controlled recording conditions and explicit motion patterns, PURE is commonly used to evaluate motion robustness in rPPG estimation.

\paragraph{UBFC-rPPG} contains 42 indoor facial video recordings in which subjects perform a time-constrained mathematical task to induce natural heart rate variations. Videos are captured using a webcam at a resolution of 640×480 pixels and 30 fps, with synchronized ground-truth PPG waveforms acquired by a CMS50E pulse oximeter. The dataset exhibits moderate appearance variation caused by mixed natural and artificial lighting, making it more challenging than PURE.

\paragraph{MMPD} is a large-scale, multi-domain rPPG dataset comprising approximately 11 hours of video recordings from 33 subjects captured using mobile phone cameras. The dataset systematically varies three key factors: skin tone (Fitzpatrick types III–VI), lighting conditions (LED high intensity, LED low intensity, incandescent, and natural lighting), and activity types (stationary, head rotation, talking, and walking). These variations introduce substantial domain shift, making MMPD a challenging benchmark for evaluating rPPG robustness under realistic conditions.
\subsection{Implementation Details}
All experiments were conducted using the open-source rPPG-Toolbox~\cite{liu2023rppg} in PyTorch. Models were trained for 30 epochs with a batch size of 4. We use $\beta = 0.05$, $\lambda_h = 0.3$, 40 Sinkhorn iterations, and $\gamma = 0.1$ as the default hyperparameter settings. For harmonic analysis, the physiological frequency range is set to $[0.7, 4.0]$ Hz. All computations were performed on an NVIDIA RTX H100 GPU.

For source-domain training, each source dataset was split into training and validation sets using an 80\%/20\% ratio. The target dataset was partitioned into two disjoint subsets: 30\% of the target data was used as unlabeled reference data to extract appearance statistics for FDA during training, while the remaining 70\% was held out as an unseen test set. Supervised learning was performed only on the original source data, and no target labels were used during training or adaptation.
\subsection{Metrics and Evaluation}
\label{sec:metrics}

To evaluate the performance of the proposed method, we adopt four standard metrics for rPPG estimation: Mean Absolute Error (MAE), Root Mean Square Error (RMSE), Mean Absolute Percentage Error (MAPE), and Pearson’s Correlation Coefficient ($r$). These metrics assess estimation accuracy and the consistency between predicted and ground-truth signals. The corresponding definitions are given as follows:

\begin{equation}
\mathrm{MAE} = \frac{1}{N} \sum_{i=1}^{N} \left| y_i - \hat{y}_i \right|
\label{eq:mae}
\end{equation}

\begin{equation}
\mathrm{RMSE} = \sqrt{ \frac{1}{N} \sum_{i=1}^{N} \left( y_i - \hat{y}_i \right)^2 }
\label{eq:rmse}
\end{equation}

\begin{equation}
\mathrm{MAPE} = \frac{100\%}{N} \sum_{i=1}^{N} \left| \frac{y_i - \hat{y}_i}{y_i} \right|
\label{eq:mape}
\end{equation}

\begin{equation}
r =
\frac{
\sum_{i=1}^{N} (y_i - \bar{y})(\hat{y}_i - \bar{\hat{y}})
}{
\sqrt{
\sum_{i=1}^{N} (y_i - \bar{y})^2
}
\sqrt{
\sum_{i=1}^{N} (\hat{y}_i - \bar{\hat{y}})^2
}
}
\label{eq:pearson}
\end{equation}

where $y_i$ and $\hat{y}_i$ denote the ground-truth and predicted values, respectively; $\bar{y}$ and $\bar{\hat{y}}$ represent their corresponding mean values; and $N$ is the total number of samples.

\subsection{Main Comparison}
\label{sec:main_comparison}

Tab.~\ref{tab:main_comparison} presents cross-dataset results of representative rPPG backbones with and without HOT. Across all evaluation scenarios, HOT consistently improves generalization under domain shift, yielding lower MAE, MAPE, and RMSE, together with higher Pearson correlation $\rho$. These gains are observed across heterogeneous architectures, including both CNN-based and Transformer-based models, indicating that the benefits of HOT are backbone-agnostic rather than architecture-specific.

Notably, the largest performance gains are observed in the PURE$\rightarrow$MMPD and UBFC-rPPG$\rightarrow$MMPD settings. As the most challenging target domain in our study, MMPD exhibits substantial illumination variation, motion interference, and recording diversity, all of which intensify appearance-driven distribution shifts. Under such severe domain discrepancy, HOT delivers more pronounced improvements, suggesting stronger robustness in preserving physiology-relevant temporal dynamics.

Fig.~\ref{fig:BlandAltman} further supports this observation from an agreement-analysis perspective. Compared with the baseline PhysNet~\cite{yu2019remote}, integrating HOT produces predictions with reduced dispersion and tighter agreement with the reference HR, as reflected by errors being more concentrated around zero in the Bland--Altman plots and estimates lying closer to the identity line in the scatter plots.

\section{Ablation Studies}
\label{sec:ablation_study}

\subsection{Low-Frequency Replacement Ratio $\beta$}
\begin{figure}[t]
  \centering
  \includegraphics[width=\linewidth]{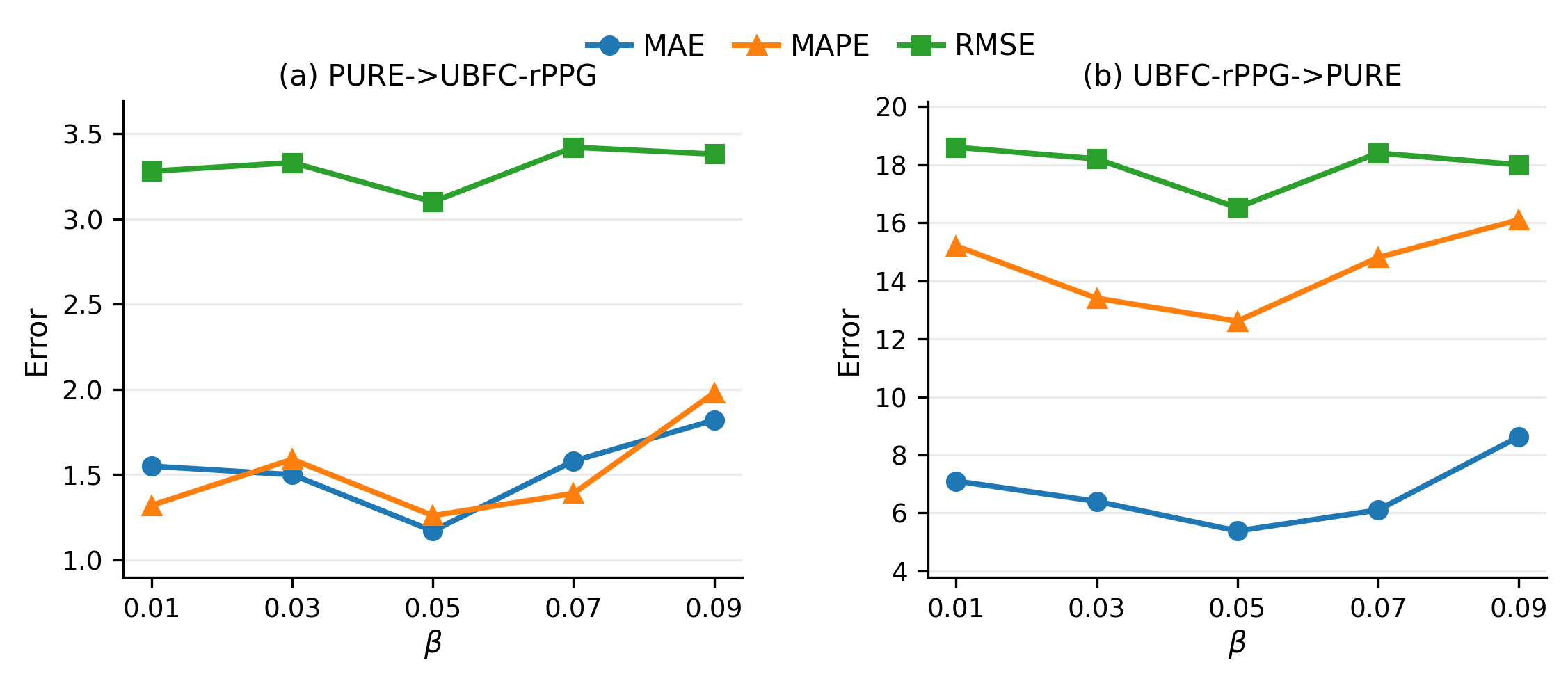}
  \caption{Effect of the low-frequency replacement ratio $\beta$ on cross-dataset performance.}
  \label{fig:beta_res}
\end{figure}

\begin{figure}[t]
  \centering
  \includegraphics[width=0.8\linewidth]{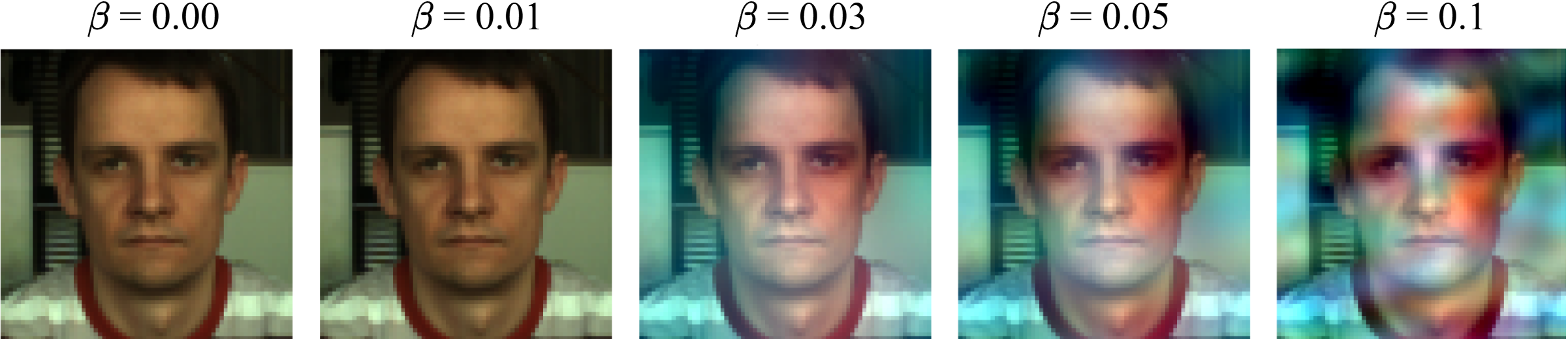}
  \caption{Visual examples of low-frequency replacement under different ratios $\beta$.}
  \label{fig:beta_vis}
\end{figure}
We evaluate PhysNet~\cite{yu2019remote} with the proposed HOT mechanism under different low-frequency replacement ratios $\beta$. As shown in Fig.~\ref{fig:beta_res}, the best overall performance is achieved at $\beta = 0.05$, which yields the lowest MAE, MAPE, and RMSE on both transfer settings: 1.17, 1.26, and 3.10 for PURE $\rightarrow$ UBFC-rPPG, and 5.38, 12.61, and 16.52 for UBFC-rPPG $\rightarrow$ PURE, respectively. This suggests that a balanced degree of low-frequency substitution effectively incorporates target-domain spectral statistics while preserving the underlying facial structure and physiological dynamics. The visual results in Fig.~\ref{fig:beta_vis} align with this observation: smaller $\beta$ results in insufficient domain alignment, whereas a larger value (e.g., $\beta = 0.1$) introduces more pronounced global appearance distortions that may interfere with subtle rPPG-related cues.
\begin{figure}[t]
  \centering
  \includegraphics[width=\linewidth]{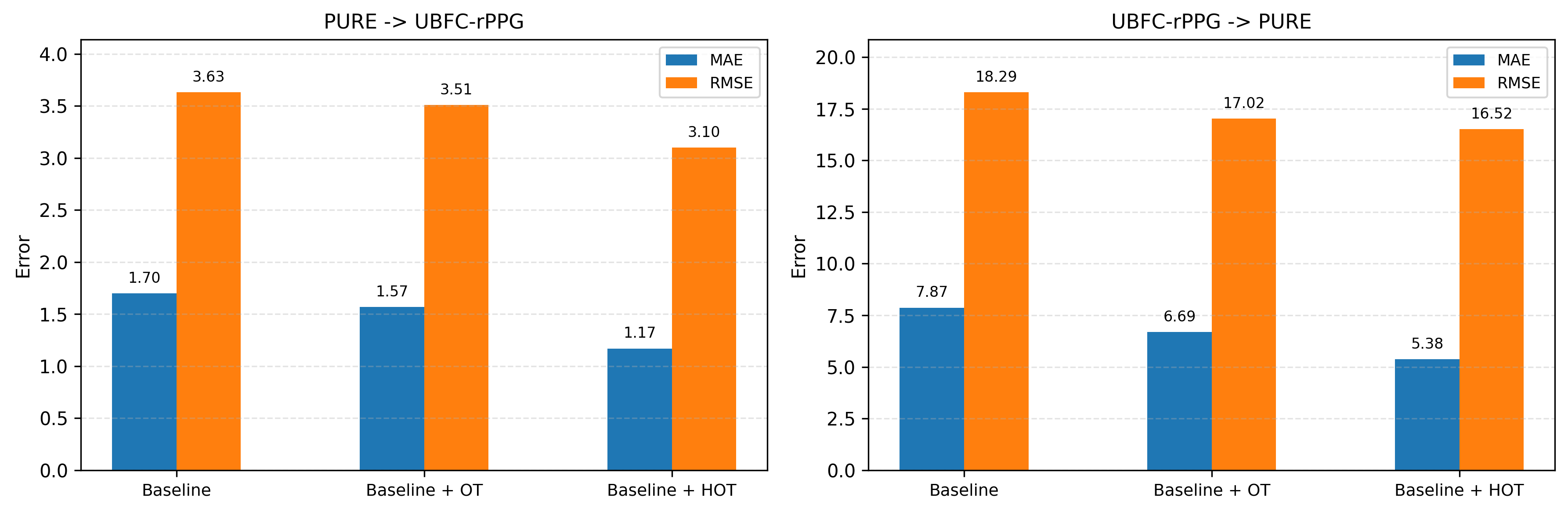}
    \caption{Performance comparison on PhysNet under progressive integration of the geometric OT loss and the harmonic-constrained OT loss into the baseline.}
  \label{fig:bar_comparison}
\end{figure}
\begin{figure}[t]
  \centering
  \includegraphics[width=0.8\linewidth]{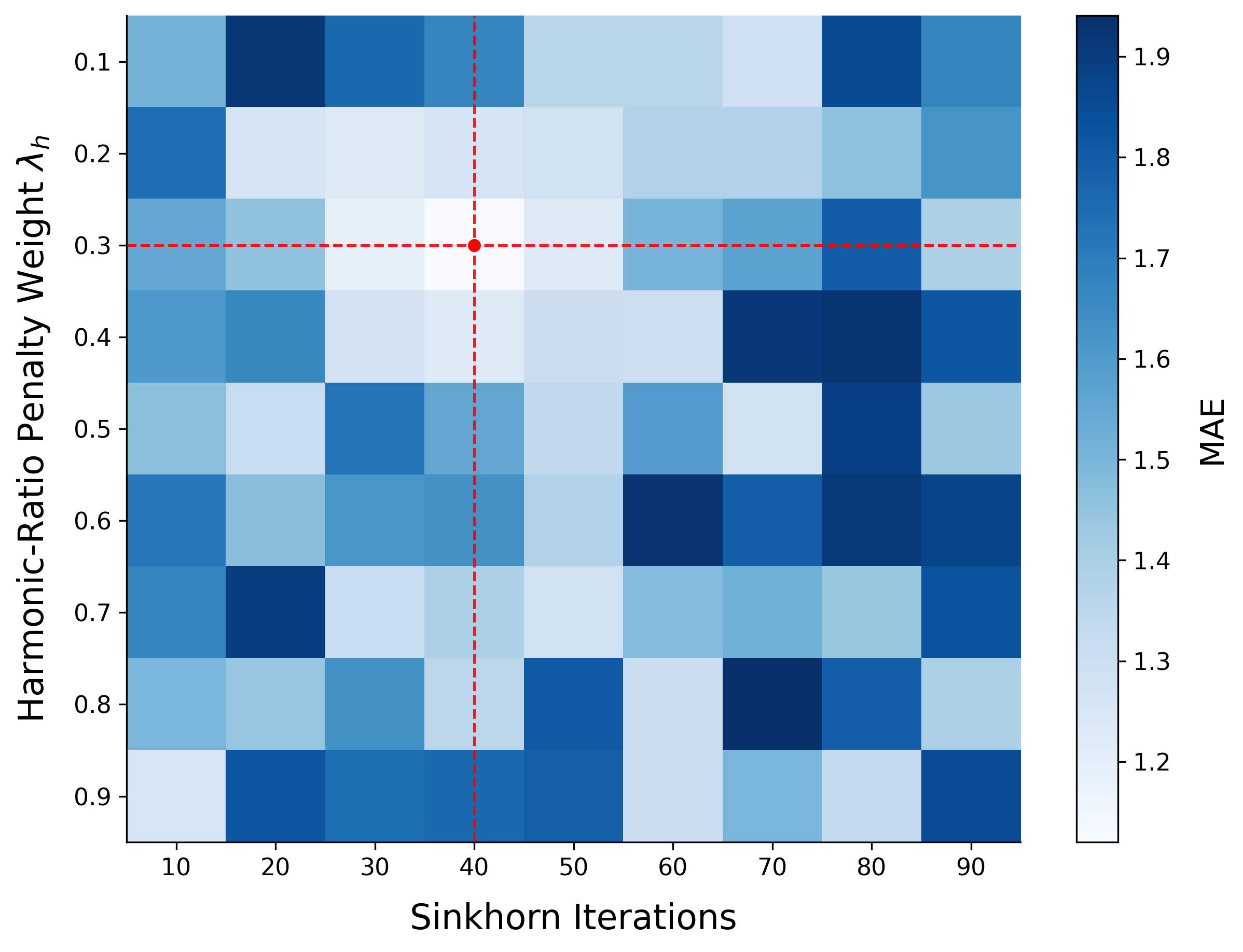}
    \caption{Heatmap of MAE across different $\lambda_h$ values and Sinkhorn iterations. Each cell corresponds to the MAE obtained under a specific hyperparameter combination. Lighter cells indicate lower error (better performance), and the red marker denotes the optimal setting.}
  \label{fig:heatmap_mae}
\end{figure}
\subsection{Effect of HOT}

To evaluate the effectiveness of the proposed harmonic-constrained loss, we adopt PhysNet~\cite{yu2019remote} as the backbone model. As illustrated in Fig.~\ref{fig:bar_comparison}, we conduct two cross-dataset adaptation experiments: PURE $\rightarrow$ UBFC-rPPG and UBFC-rPPG $\rightarrow$ PURE. The objective is to examine whether the harmonic constraint can guide the Sinkhorn solver to produce a transport plan that aligns tokens sharing consistent physiological context across domains. The results demonstrate improved cross-dataset performance, indicating more reliable physiological correspondence and enhanced robustness under domain shift.
Fig.~\ref{fig:heatmap_mae} presents a sensitivity analysis of the harmonic-ratio penalty weight $\lambda_h$ and the number of Sinkhorn iterations under the PURE $\rightarrow$ UBFC-rPPG setting. The best configuration is obtained at $\lambda_h = 0.3$ with 40 iterations, which we use as the default setting in our experiments. This result indicates that a moderate harmonic penalty provides effective regularization for the transport plan without over-constraining it. We also observe that increasing the number of iterations beyond 40 incurs higher computational cost, while yielding no further performance improvement.
\begin{table}[t]
\centering
\footnotesize
\setlength{\tabcolsep}{3.5pt}
\resizebox{\columnwidth}{!}{%
\begin{tabular}{lcccc}
\toprule
Method & Train Time $\downarrow$ & Peak Memory $\downarrow$ & Latency $\downarrow$ & Throughput $\uparrow$ \\
       & (ms/iter) & (GB) & (ms) & (Kfps) \\
\midrule
PhysNet w/o HOT & 391.3 & 1.27 & 1.92 & 66.75 \\
PhysNet w/ HOT  & 434.8 & 2.52 & 1.89 & 67.82 \\
\midrule
Relative Change 
& \textcolor{red!75!black}{\textbf{+11.1\%}}
& \textcolor{red!75!black}{\textbf{+98.4\%}}
& \textcolor{green!50!black}{\textbf{-1.6\%}}
& \textcolor{green!50!black}{\textbf{+1.6\%}} \\
\bottomrule
\end{tabular}%
}
\caption{Computational overhead of HOT on PhysNet~\cite{yu2019remote}. Training efficiency is measured by per-iteration training time and peak GPU memory. Inference efficiency is measured by forward-only latency and throughput. Lower is better for Train Time, Peak Memory, and Latency; higher is better for Throughput.}
\label{tab:hot_overhead}
\end{table}
\subsection{Computational Overhead Analysis}
We report the computational overhead of HOT on top of PhysNet~\cite{yu2019remote} in Table~\ref{tab:hot_overhead}. HOT increases the training time from 391.3 to 434.8 ms/iter and the peak GPU memory from 1.27 to 2.52 GB. In contrast, the inference efficiency remains nearly unchanged, with comparable latency (1.89 vs.\ 1.92 ms) and similar throughput (67.82 vs.\ 66.75 Kfps), since HOT is only applied during training and is not involved at inference time. These results show that HOT mainly introduces additional cost during training, while leaving inference efficiency essentially unchanged.
\section{Conclusion}
\label{sec:Conclusion}
In this work, we investigated the problem of cross-domain robustness in rPPG by explicitly modeling appearance variation and enforcing physiologically consistent alignment during training. We introduced FDA as a data-level strategy to account for domain-dependent appearance characteristics in rPPG videos, encouraging models to learn representations invariant to such variations while preserving cardiac-induced signals. To complement FDA-based training, we proposed HOT as a regularization loss that leverages the harmonic property of cardiac signals to guide alignment between original and appearance-shifted representations. Extensive cross-dataset evaluations demonstrate that the proposed FDA + HOT framework consistently improves the robustness and generalization of rPPG models under domain shift, providing an effective solution for cross-domain rPPG estimation. Despite the promising results, this study has several limitations. The current FDA pipeline relies on unlabeled target-domain reference samples during training, which may restrict applicability when such reference data is unavailable. In addition, HOT introduces extra hyperparameters and computational cost due to Sinkhorn-based alignment. Future work will explore more flexible variants that reduce dependence on target-domain reference data and improve computational efficiency, together with broader validation under more diverse real-world conditions.
\section*{Acknowledgment}
This work was supported by VinUniversity’s Seed Grant Program under project VUNI.2425.EME.005 and the National Foundation for Science and Technology Development (NAFOSTED) through Project IZVSZ2-229539 (2025–2027). The authors would like to sincerely thank the Vietnam Young Talent Support Fund, Tan Hiep Phat Trading – Service Co., Ltd., and the Ben Dam Me Award Fund for their valuable support and encouragement of this work.  
\label{sec:acknowledgment}
{
    \small
    \bibliographystyle{ieeenat_fullname}
    \bibliography{main}
}

\end{document}